\documentclass[conference]{IEEEtran}
\IEEEoverridecommandlockouts
\ifCLASSINFOpdf
\else
\fi
\usepackage{color,soul}
\usepackage{graphicx}
\usepackage{multirow}
\usepackage{pifont}
\usepackage{booktabs}
\usepackage{pifont} 
\usepackage{colortbl} 
\usepackage{authblk}
\usepackage{url}
\usepackage{footmisc}
\usepackage{hyperref}
\usepackage{amsmath}
\usepackage{flushend}
\usepackage[table]{xcolor}
\usepackage[comma, sort&compress,square,numbers]{natbib}
\newcommand{\cmark}{\textcolor{blue}{\ding{51}}} 
\newcommand{\xmark}{\textcolor{red}{\ding{55}}} 
\begin{document}
%
\title{BVI-CR: A Multi-View Human Dataset for Volumetric Video Compression\\

\thanks{The authors appreciate the funding from the Innovate UK MyWorld Collaborative Research \& Development Scheme (10038316).}}
%
%
%
\author[1]{Ge Gao}
\author[1]{Adrian Azzarelli}
\author[1]{Ho Man Kwan}
\author[1]{Nantheera Anantrasirichai}
\author[1]{Fan Zhang}
\author[2]{\\Oliver Moolan-Feroze}
\author[1]{David Bull}
\affil[1]{\textit{Bristol Vision Institute, University of Bristol, Bristol, BS1 5DD, United Kingdom}}
\affil[1]{\textit {\{ge1.gao, adrian.azzarelli, hm.kwan, n.anantrasirichai, fan.zhang, dave.bull\}@bristol.ac.uk}}
\affil[2]{\textit{Condense Reality Ltd, 1 Canon's Road, Bristol, BS1 5TX}}
\affil[2]{\textit{ollie@condensereality.com}}

\maketitle
\begin{abstract}
The advances in immersive technologies and 3D reconstruction have enabled the creation of digital replicas of real-world objects and environments with fine details. These processes generate vast amounts of 3D data, requiring more efficient compression methods to satisfy the memory and bandwidth constraints associated with data storage and transmission. However, the development and validation of efficient 3D data compression methods are constrained by the lack of comprehensive and high-quality volumetric video datasets, which typically require much more effort to acquire and consume increased resources compared to 2D image and video databases. To bridge this gap, we present an open multi-view volumetric human dataset, denoted BVI-CR, which contains 18 multi-view RGB-D captures and their corresponding textured polygonal meshes, depicting a range of diverse human actions. Each video sequence contains 10 views in 1080p resolution with durations between 10-15 seconds at 30FPS. Using BVI-CR, we benchmarked three conventional and neural coordinate-based multi-view video compression methods, following the MPEG MIV Common Test Conditions, and reported their rate quality performance based on various quality metrics. The results show the great potential of neural representation based methods in volumetric video compression compared to conventional video coding methods (with an up to 38\% average coding gain in PSNR). This dataset provides a development and validation platform for a variety of tasks including volumetric reconstruction, compression, and quality assessment. The database will be shared publicly at \url{https://github.com/fan-aaron-zhang/bvi-cr}.
\end{abstract}

\begin{IEEEkeywords}
Volumetric video databases, BVI-CR, textured mesh, point cloud, volumetric video compression
\end{IEEEkeywords}

\IEEEpeerreviewmaketitle

\section{Introduction}
Volumetric video, also known as free-viewpoint video, is a form of immersive media that enables consumers  to explore real content in 3-D from arbitrary angles in space. This offers the potential for significantly enhanced interactive experiences compared to traditional video content. Recently, advances in capture, display and rendering technologies have enhanced the volumetric video production pipeline, making volumetric video more viable for use in various virtual reality (VR) and augmented reality (AR) applications, such as livestreaming concerts~\cite{onderdijk2023concert} or sports~\cite{baia2022video}, immersive conferencing~\cite{gunkel2018virtual} and XR museum exhibits~\cite{lee2020experiencing}.

Volumetric data compression is essential to enable high-quality immersive experiences, particularly when delivered over bandlimited networks. In the past decade, volumetric video compression has attracted increased attention in both academia and industry sectors, leading to the development of several MPEG standards such as G-PCC/V-PCC~\cite{schwarz2018emerging} for point clouds compression, and MV-HEVC~\cite{hannuksela2015overview} and MIV~\cite{boyce2021mpeg} for multi-view video compression. 
More recently, numerous implicit, coordinate-based neural representations (INRs) have also been proposed for for both  conventional  and volumetric video processing ~\cite{mildenhall2021nerf,xie2022neural,pumarola2021d,azzarelli2023waveplanes,gao2024pnvc,kwan2024nvrc} and compression~\cite{zhu2023implicit,zhang2024efficient,kwan2024immersive}. These offer significant potential to compete, and even outperform, conventional approaches.

\begin{figure}[t!]
    \centering
    \includegraphics[width=\linewidth]{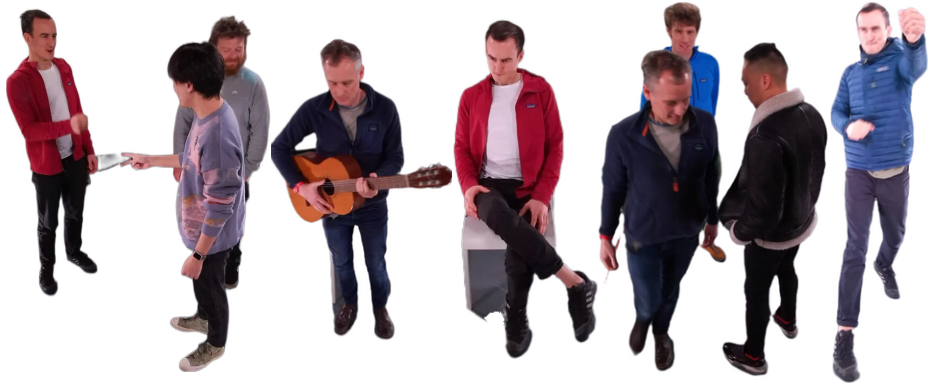}
    \caption{Snapshots of exemplary scenes from the BVI-CR dataset, comprising both single person actions and multiple person interactions.}
    \label{fig:snapshots}
\end{figure}

\begin{figure*}[t]
    \centering
    \includegraphics[width=\linewidth]{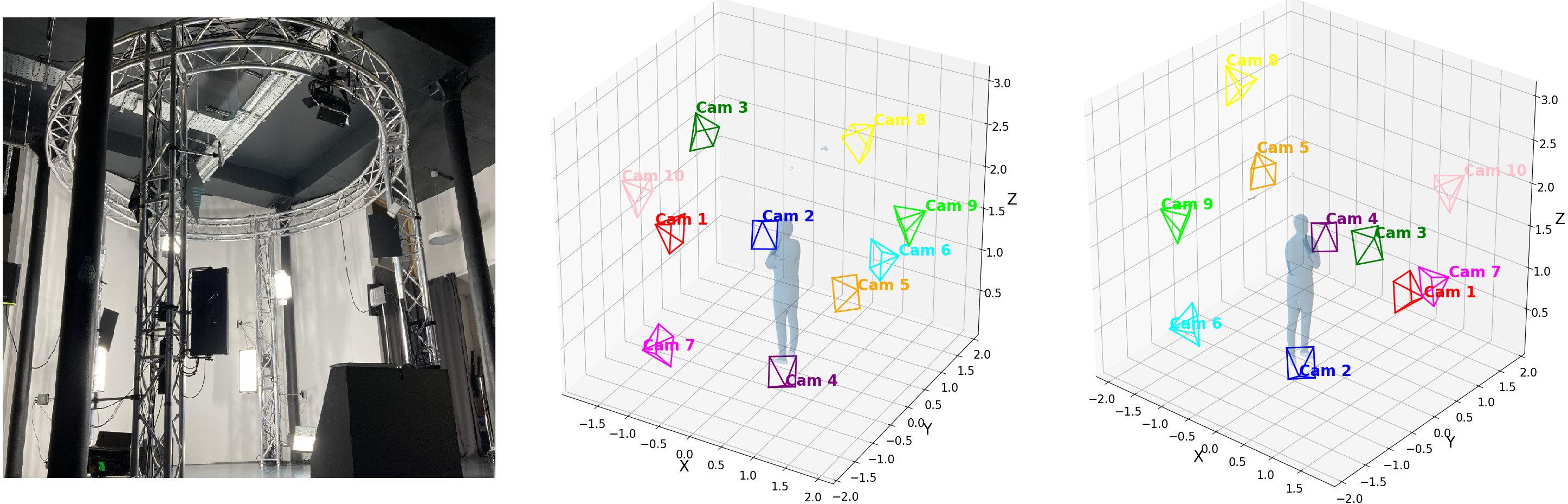}
    \caption{(\textbf{Left}) the Metaverse Studio used for content capture. (\textbf{Right}) Illustration of the camera arrangement from two different angles.}
    \label{fig:capture-rig}
\end{figure*}

As with other video processing tasks, high-quality and diverse video datasets are essential to facilitate the development and evaluation of neural representations and compression algorithms. However, existing volumetric datasets~\cite{xu2017owlii,ionescu2013human3,Joo_2017_TPAMI,Yu_2020_CVPR,habermann2021,gautier2023uvg,zerman2020textured,peng2021neural,li2021learn,cheng2022generalizable} are either not designed for compression applications or lack diversity in their video content and/or format. In this paper, we specifically focus on human-object interaction and introduce a new multi-view video dataset, BVI-CR\footnote{Here BVI and CR stand for the organizations of authors, Bristol Vision Institute and Condense Reality Ltd, respectively.}, featuring 18 video sequences that capture a variety people, object types and actions. Each sequence comprises 10 synchronized HD-resolution videos, each between 10-15 seconds long (300-450 frames). For each sequence, we also provide metadata including the corresponding depth maps, background masks and textured meshes generated based on their multi-view content. Sample frames of a multi-view sequence in this database are shown in Fig. \ref{fig:snapshots}. To demonstrate its application for volumetric video compression, we have benchmarked one conventional and two INR-based immersive video codecs on the BVI-CR database, showing their rate quality performance evaluated using three different metrics. It should be noted that our database can be used for other 3D video tasks such as volumetric scene reconstruction, neural rendering, and quality assessment. 


\section{The BVI-CR dataset}
This section describes the acquisition configuration used to capture the multi-view video sequences, and presents the main features of BVI-CR dataset.

\subsection{Capture Configuration}

\noindent\textbf{Volumetric Video Capture Studio.} The BVI-CR content was captured using a volumetric video capture studio developed by Condense Reality (CR). The facility is shown in Fig.~\ref{fig:capture-rig} (\textbf{Left}) and  consists of ten Microsoft Azure cameras that capture $\text{2560} \times \text{1440}$ resolution RGB video at 30FPS. Each camera is equipped with a Time-of-Flight (ToF) depth sensor producing $\text{640} \times \text{576}$ resolution depth images, synchronized at the same frame rate. The Azure units are synchronized, ensuring temporally consistent multi-view capture.


The RGB and depth sensors are configured on a hemisphere surrounding the central capture area, with a diameter of 2.5m. The ten pairs of cameras and sensors are placed symmetrically around the central area, with fixed baselines as defined by the product specification. As shown in Fig.~\ref{fig:capture-rig}, the cameras are positioned to optimize scene coverage, where one pair is in front of the actor, one is behind, and four pairs are placed in a $2 \times 2$ grid formation on both sides of the actor. This configuration supports a wide range of angles and depths, encapsulating the entire stage volume and providing diverse stereo disparity to improve volumetric data quality.

\begin{table*}[t!]
    \centering
    \caption{Key features of existing open-sourced volumetric human performance datasets and the proposed BVI-CR dataset. Here, the superscript * means low-resolution VGA cameras.}
    \label{tab:comparison}
    \resizebox{0.95\textwidth}{!}{\begin{tabular}{rccccccccc}
        \hline
        \hline
        \rowcolor{lightgray}
        Dataset & \#seqs. & FPS & \#frames & \#views & Multi-view RGB & Multi-view Depth & 3D Meshes & Multi-person \\ 
        \hline 
        \hline
        Owlii~\cite{xu2017owlii} & 4 & 30 & 600 & - & \xmark & \xmark & \cmark & \xmark \\ \hline
        HUMBI (Body)~\cite{Yu_2020_CVPR} & 772 & 60 & 26M & 107 & \cmark & \xmark & \cmark & \xmark \\ \hline
        Neural Actor~\cite{habermann2021} & 8 & - & 20,000 & 79-86 & \cmark & \cmark & \cmark & \xmark \\ \hline
        UVG-VPC~\cite{gautier2023uvg} & 12 & 25 & 3,000 & 96 & \xmark & \xmark & \cmark & \xmark \\ \hline
        vsenseVVDB2~\cite{zerman2020textured} & 3 & 30 & 149--1830 & 12/60 & \cmark & \xmark & \xmark & \xmark \\ \hline
        ZJU-MoCap~\cite{peng2021neural} & 9 & 21 & 15M & 42 & \cmark & \cmark & \cmark & \cmark \\ \hline
        AIST++~\cite{li2021learn} & 1408 & - & 10.1M & 9 & \xmark & \xmark & \cmark & \xmark \\ \hline
        THUMan 4.0~\cite{deepcloth_su2022} & - & - & 10,000 & 24 & \cmark & \cmark & \cmark & \xmark \\ \hline
        GeneBody~\cite{cheng2022generalizable} & 16 & - & 2.95M & 48 & \cmark & \xmark & \cmark & \cmark \\ \hline
        \textbf{BVI-CR (Ours)} & 18 & 25 & 7,200 & 10 & \cmark & \cmark & \cmark & \cmark \\ \hline
    \end{tabular}}
\end{table*}

\begin{figure*}[t!]
    \centering
    \includegraphics[width=\linewidth]{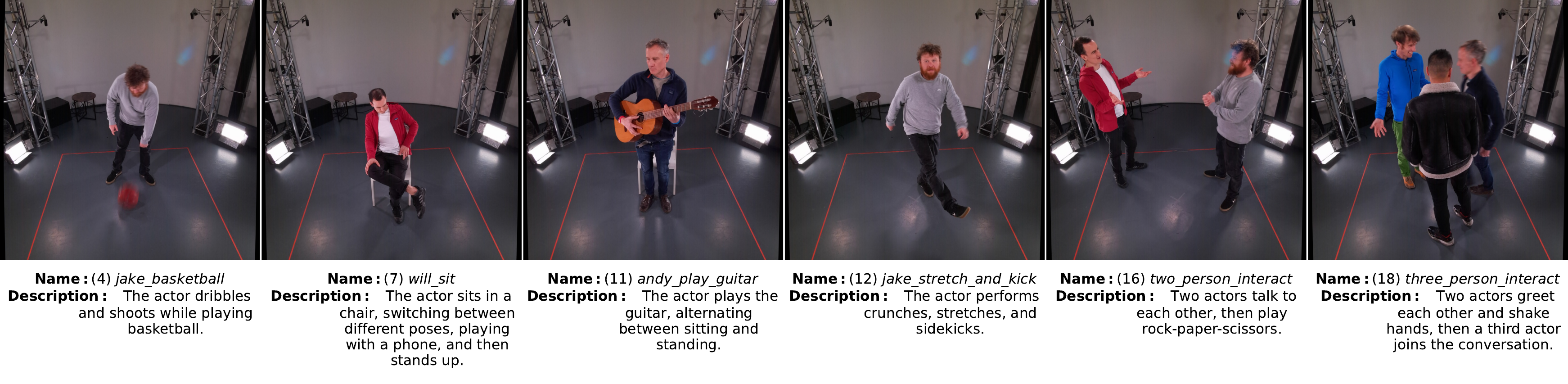}
        \vspace{-20pt}
    \caption{Selected pose examples with the sequence name, sequence index, and descriptions to the complete actions performed.}
    \label{fig:snapshots-action}
\end{figure*}

\noindent\textbf{Data Capture.} At the beginning of each capture session, geometric calibration is performed on all cameras and sensors to accurately recover 3D pose data. The calibration process defines a unified spatial coordinate system for the multi-camera rig, where a common origin and orientation are determined using a reference camera and a standardized rotation matrixrespect to all other cameras are positioned and oriented. This ensures that all cameras are aligned in the 3D space, allowing for robust real-time generation of volumetric content  using a fusion of RGB and depth data, during capture. Once calibration is accomplished, the human actors then perform their scripted actions within the capture volume. This setup process minimizes depth data misalignment and improves the accuracy of the resulting volumetric reconstructions and textures.

\noindent\textbf{Volumetric Data Generation.} The volumetric data generation is  based on ~\cite{newcombe2015dynamicfusion}, and adapted for mutli-camera recording. Given a a set of calibrated RGBD cameras, we can generate a series of volumetric frames represented by 3D polygonal meshes and associated textures. For each set of input RGB and depth frames, we first fuse the data into a 3D Truncated Signed Distance Function (TSDF) volume. The TSDF represents the surface of the objects being recorded at its zero level set. As new frames are created they are integrated into the initial volume via a non-rigid registration which allows us to reduce noise, fill holes and conceal occlusion artifacts. To generate an output at a given time point, we apply registration and extract a mesh using the marching cubes algorithm~\cite{lorensen1998marching}. We finally create a texture to the output by computing a triangle-based UV mapping, which generates a compact texture image with associated per-triangle UV coordinates.

\subsection{Video Sequences and Associated Data}
Based on the configuration and capture pipeline described above, we have collected 18 sequences featuring 7 actors/actresses performing a variety of actions and poses. These include 15 single-human scenes and 3 multi-human interactions, involving 2 to 3 subjects. Each sequence is 10-15 seconds long and composed of 300-450 frames. The actions were designed to (i) encompass a range of movements, from basic to complex, involving both upper and lower body, with particular attention to limb movements, and (ii) introduce occlusions through human-object interactions, such as sitting on a chair or bouncing a basketball. As shown in Fig.~\ref{fig:snapshots-action}, the actions include running, sitting down/standing up, dribbling, punching, and kicking, sometimes with facial obstructions due to hats or sunglasses. Multi-person scenarios include talking, whispering, and shaking hands. These scenes reflect common human activities and present various challenges for multi-view and volumetric reconstruction, rendering, and compression algorithms.

In conclusion, each scene comprises: (i) 10-view RGB videos with $1920 \times 1080$ resolution, (ii) 10-view depth videos with $512 \times 424$ resolution, (iii) camera positions, orientations and intrinsics for RGB and depth cameras, (iv) masks for background matting that were generated by the pretrained FCN-ResNet50~\cite{long2015fully}, which was empirically found to give the most consistent performance, and v) high-quality textured 3D mesh sequences provided in OBJ, PNG, and MTL formats.

\section{Multiview Compression Benchmarking}
While the BVI-CR database can support multiple volumetric video processing tasks, we focus here on solely on video compression. We benchmarked three state-of-the-art, open-source multi-view video codecs including: (i) an MPEG standard codec, TMIV~\cite{boyce2021mpeg} using the main anchor configuration, where VVenC~\cite{wieckowski2021vvenc} in Random Access mode is used to encode both texture and depth information; (ii) two implicit neural representation (INR)-based codecs, MV-HiNeRV~\cite{kwan2024immersive} and MV-IERV~\cite{zhu2023implicit}. Compression performance was evaluated based on the (masked) 10 camera views with associated RGB ground truth, together with 4 additionally sampled front-facing views used to assess the reconstruction quality of synthesized views. The raw RGB and depth frames are converted to videos in YUV 4:2:0 format based on the BT.709 standard. For MV-IERV, we select the front-facing view as the base view that is compressed by an HEVC/H.265 codec (x.265, the default \textit{veryfast} configuration). All synthesized views are rendered using the TMIV default view synthesizer~\cite{boyce2021mpeg} given the decoded texture attributes (of ten source views), the camera view parameters and the viewport parameters. The fidelity of the compressed source views is evaluated using
PSNR, SSIM and immersive video PSNR (IV-PSNR)~\cite{dziembowski2022iv} following~\cite{boyce2021mpeg}. Here, IV-PSNR extends the traditional PSNR metric for immersive videos by accounting for spherical geometry and projection distortions, applying weighted error calculations based on pixel importance, and incorporating perceptual factors to align with human visual sensitivity for more accurate quality assessment. Bj{\o}ntegaard Delta Rate (BD-rate)~\cite{bdrate} (with TMIV being the anchor) is then employed to measure the coding performance. The INR-based models MV-HiNeRV and MV-IERV are optimized following the training methodology described in their original papers~\cite{kwan2024immersive,zhu2023implicit}.

\begin{figure*}[t!]
    \centering
    \includegraphics[width=0.9\linewidth]{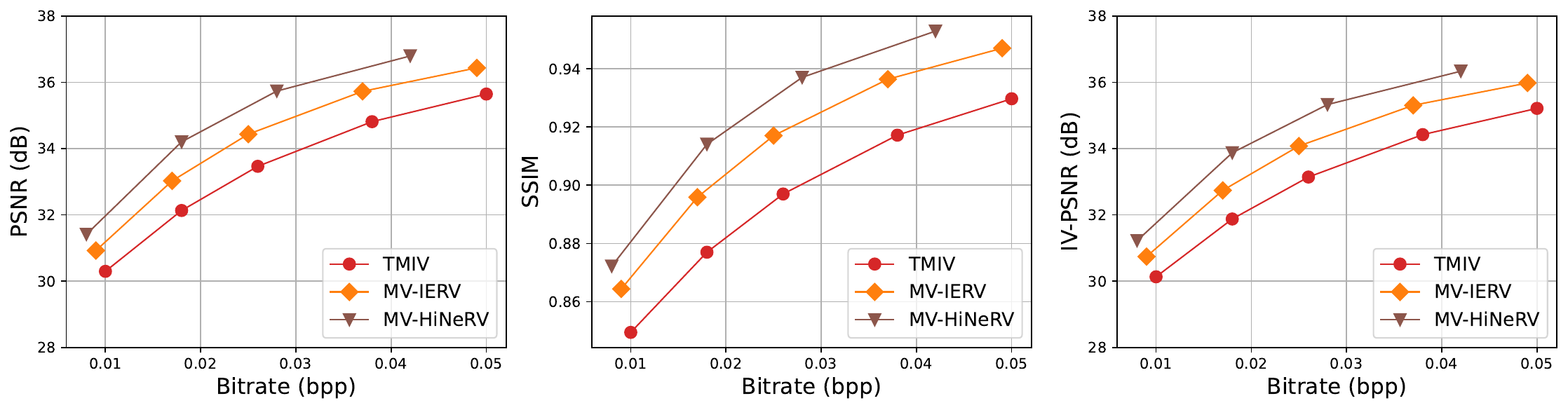}
    \vspace{-5pt}
    \caption{Rate-distortion performance comparison of selected baselines, with distortion measured by PSNR, SSIM, and IV-PSNR, respectively.}
    \label{fig:rd-plot}
\end{figure*}

\begin{figure}[t!]
    \centering
    \includegraphics[width=\linewidth]{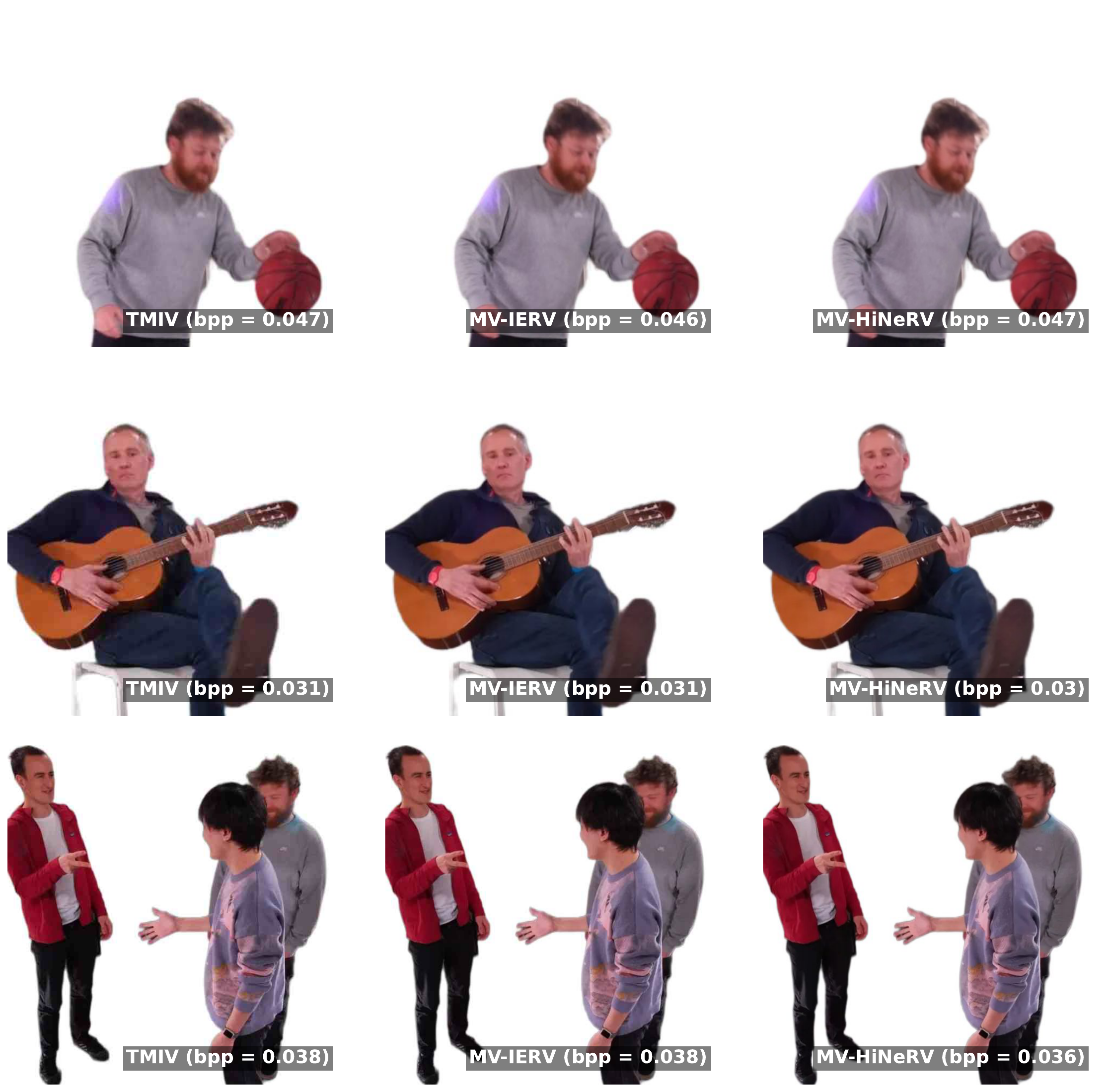}
    \caption{Qualitative comparison of synthesized views (cropped) for 3 randomly sampled frames.}
    \label{fig:visual-comparison}
\end{figure}

Fig.~\ref{fig:rd-plot} illustrates the rate-distortion performance of the three benchmarked volumetric video codecs based on three quality metrics. It can be observed that MV-HiNeRV attains the best overall performance for all metrics and sequences measured. This is further demonstrated by an average of 38.5\% and 34.3\% reduction in BD-rate, measured by PSNR and IV-PSNR, respectively, compared to TMIV anchor, as shown in Table~\ref{tab:bd-rate}. In comparison, the MV-IERV codec outperforms TMIV by an average of 29.5\% and 26.9\% in terms of BD-rate, measured by PSNR and IV-PSNR, respectively. The same performance holds for perceptual quality assessment on the synthesized viewports. As it can be seen from Fig.~\ref{fig:visual-comparison}, the scenes rendered from MV-HiNeRV's outputs contain fewer visual artifacts as in TMIV and MV-IERV. However, the relative performance gain of both INR-based codecs is reduced in the multi-human scenes, persumably due to the increased complexity of the scenes and the larger number of views that are occluded. 

\begin{table}[ht]
\centering
\caption{BD-rates (\%) of the INR-based codecs w.r.t TMIV. Negative BD-rate values indicate better compression efficiency. Here, * denotes the scenes involving multiple actors.}
\label{tab:bd-rate}
\resizebox{0.98\linewidth}{!}{\begin{tabular}{r|ccc|ccc}
\toprule
  & \multicolumn{3}{c|}{\textbf{MV-HiNeRV}} & \multicolumn{3}{c}{\textbf{MV-IERV}} \\
\cmidrule{2-7}
\textbf{Seq.}& PSNR & SSIM & IV-PSNR & PSNR & SSIM & IV-PSNR \\
\midrule
1 & -41.2 & -43.7 & -36.8 & -31.8 & -33.4 & -28.9 \\
2 & -39.8 & -42.2 & -45.5 & -30.6 & -32.1 & -37.7 \\
3 & -52.5 & -54.9 & -47.9 & -42.9 & -44.3 & -39.8 \\
4 & -28.3 & -30.8 & -34.1 & -19.3 & -21.9 & -26.5 \\
5 & -40.7 & -43.1 & -26.3 & -31.4 & -32.9 & -18.4 \\
6 & -37.6 & -40.2 & -43.5 & -28.7 & -30.4 & -35.0 \\
7 & -53.1 & -55.4 & -48.4 & -43.4 & -44.7 & -40.2 \\
8 & -29.2 & -31.7 & -24.9 & -20.1 & -22.7 & -17.3 \\
9 & -51.8 & -54.2 & -47.3 & -42.3 & -43.7 & -39.2 \\
10 & -38.9 & -41.4 & -34.6 & -29.9 & -31.5 & -27.1 \\
11 & -40.3 & -42.7 & -35.9 & -31.1 & -32.6 & -28.1 \\
12 & -27.2 & -29.8 & -33.1 & -18.4 & -21.1 & -25.7 \\
13 & -52.8 & -55.1 & -38.1 & -43.2 & -44.5 & -30.0 \\
14 & -39.5 & -42.0 & -45.2 & -30.4 & -31.9 & -37.5 \\
15 & -41.5 & -43.9 & -27.0 & -32.0 & -33.5 & -19.0 \\
16* & -24.8 & -27.6 & -21.0 & -16.5 & -19.4 & -14.0 \\
17* & -23.9 & -26.7 & -20.2 & -15.8 & -18.7 & -13.3 \\
18* & -25.2 & -28.0 & -21.4 & -16.9 & -19.7 & -14.3 \\
\midrule
Overall & -38.5 & -41.1 & -34.3 & -29.5 & -31.2 & -26.9 \\
\bottomrule
\end{tabular}}
\end{table}

\section{Conclusion}
In this paper, we present an open 4D volumetric dynamic human dataset, BVI-CR that features 15 sequences of individual human performances and 3 multi-human sequences capturing interactions between 2 to 3 people.  Each scene comprises the raw RGB and depth scans and the high-quality textured meshes. These sequences include diverse actions including fast motions, complex deformations, and occlusions, designed to offer both challenging training conditions for developing more robust algorithms and harsher scenarios for more rigorous evaluation of volumetric visual tasks. We further benchmark two INR-based immersive video codecs against the MPEG test model TMIV, demonstrating the superior performance and promising potential of this line of work. We believe that the BVI-CR dataset will serve as a valuable resource for researchers, fostering advances in immersive visual media technologies. For future work, we plan to extend the dataset by incorporating a wider variety of actions and actor appearances, and also include audio recordings.

\small
\bibliographystyle{IEEEtran}
\bibliography{IEEEexample}

%








\end{document}